\documentclass[10pt,twocolumn,letterpaper]{article}

\usepackage{cvpr}
\usepackage{times}
\usepackage{epsfig}
\usepackage{graphicx}
\usepackage{amsmath}
\usepackage{amssymb}
\usepackage{breqn}
\usepackage{tabularx}
\usepackage{multirow}

\usepackage{lineno}

\usepackage{epstopdf}

\usepackage[pagebackref=true,breaklinks=true,letterpaper=true,colorlinks,bookmarks=false]{hyperref}

\cvprfinalcopy 

\ifcvprfinal\fi
\begin{document}

\title{Mind the Class Weight Bias: Weighted Maximum Mean Discrepancy\\for Unsupervised Domain Adaptation}

\author{Hongliang Yan$^{1}$,  Yukang Ding$^{1}$,  Peihua Li$^2$,  Qilong Wang$^{2}$,  Yong Xu$^{3}$,  Wangmeng Zuo$^{1,}$\thanks{Corresponding author.}\\
$^1$School of Computer Science and Technology, Harbin Institute of Technology, Harbin, China\\
$^2$School of Information and Communication Engineering, Dalian University of Technology, Dalian, China\\
$^3$Bio-Computing Research Center, Shenzhen Graduate School, Harbin Institute of Technology, Shenzhen, China\\
{\tt\small yanhl@hit.edu.cn, dingyukang921@163.com, peihuali@dlut.edu.cn,}\\
{\tt\small qlwang@mail.dlut.edu.cn, laterfall@hitsz.edu.cn, wmzuo@hit.edu.cn}
}

\maketitle
\thispagestyle{empty}

\begin{abstract}
   In domain adaptation, maximum mean discrepancy (MMD) has been widely adopted as a discrepancy metric between the distributions of source and target domains.
   However, existing MMD-based domain adaptation methods generally ignore the changes of class prior distributions, i.e., class weight bias across domains. This remains an open problem but ubiquitous for domain adaptation, which can be caused by changes in sample selection criteria and application scenarios. We show that MMD cannot account for class weight bias and results in degraded domain adaptation performance. To address this issue, a weighted MMD model is proposed in this paper. Specifically, we introduce class-specific auxiliary weights into the original MMD for exploiting the class prior probability on source and target domains, whose challenge lies in the fact that the class label in target domain is unavailable. To account for it, our proposed weighted MMD model is defined by introducing an auxiliary weight for each class in the source domain, and a classification EM algorithm is suggested by alternating between assigning the pseudo-labels, estimating auxiliary weights and updating model parameters. Extensive experiments demonstrate the superiority of our weighted MMD over conventional MMD for domain adaptation. 

\end{abstract}
\section{Introduction}\label{section1}
Deep convolutional neural networks (CNNs) have achieved great success in various computer vision tasks such as image classification~\cite{krizhevsky2012imagenet}, object detection~\cite{girshick2014rich} and semantic segmentation \cite{long2015fully}.
\begin{figure}[!htp]
\begin{center}
\includegraphics[width=0.95\linewidth,height = 4.0cm]{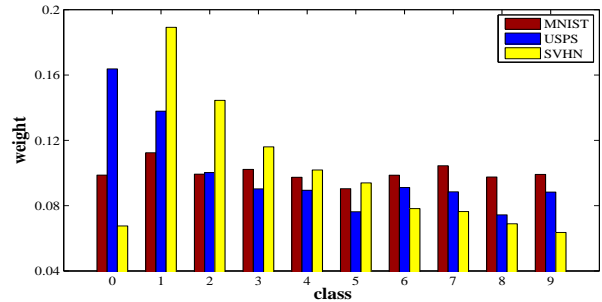}
\end{center}
   \caption{Class prior distributions of three domains for digit recognition. As is shown, class bias exists across domains. It is natural to see that the class weight of 0 and 1 are relatively high in postal service (\emph{USPS}), and the class weight of 1 and 2 are relatively high in house numbers (\emph{SVHN}).}
   \label{fig1}
\end{figure}
Besides the inspiring progress in model and learning, the achievement of CNN is undoubtedly attributed to the availability of massive labeled datasets. For a CNN trained on large scale datasets~\cite{donahue2014decaf}, while the lower layers of features are safely transferable, the learned features gradually moves from general to specific along the network \cite{yosinski2014transferable}.
When the source and target tasks are significantly diverse, the CNN pretrained on the source task may not generalize well to the target task. Such scenario leads to an emerging topic to transfer the CNN from the source task to the target task with the enhanced and discriminative representation~\cite{azizpour2015generic}. In this work, we study a special type of transfer learning task, \ie, domain adaptation (DA)~\cite{pan2010survey}.
\begin{figure*}[!htb]
\begin{center}
\includegraphics[width=0.9\linewidth ,height = 6.0cm]{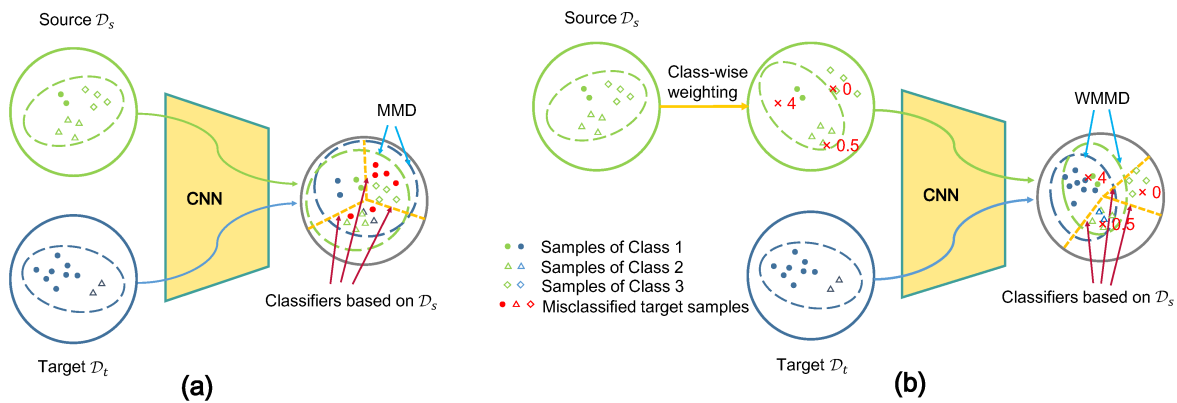}
\end{center}
   \caption{Results of minimizing MMD and WMMD regularizer under class weight bias are illustrated in (a) and (b), respectively. Minimizing MMD preserves the class weights in source domain and thus the target samples will be wrongly estimated, as indicated by yellow samples. On the contrary, the proposed weighted MMD removes the effect of class bias by first reweighting source data.}
   \label{fig2}
\end{figure*}

One of the most fruitful lines for DA is MMD-based method~\cite{long2013transfer, pan2011domain, baktashmotlagh2016distribution, zhong2009cross, tzeng2014deep}. Despite the great success achieved, existing ones generally ignore the changes of class prior distributions, dubbed by class weight bias. It is ubiquitous for domain adaptation and can be caused by changes in sample selection criteria~\cite{huang2006correcting} and application scenarios \cite{ming2015unsupervised}. As shown in Fig.~\ref{fig1}, the class prior distributions (\ie, class weights) vary with domains for digit recognition. Moreover, a special case of class weight bias is the imbalanced cross-domain data problem~\cite{ming2015unsupervised} where several classes in source domain do not appear in target domain, as shown in Fig.~\ref{fig2}. The Closest Common Space Learning (CCSL) method in \cite{ming2015unsupervised} is suggested for imbalanced and multiple cross-domain visual classification. However, CCSL just combines conventional MMD with domain-dependent MMD without explicitly considering class weight bias.

For MMD-based methods, the ignorance of class weight bias can deteriorate the domain adaptation performance. In the case of class weight bias, the MMD can be minimized by either learning domain-invariant representation or preserving the class weights in source domain. As illustrated in Fig.~\ref{fig2} (a), it is unreasonable for domain adaptation to require that the class weights in target domain should keep the same as those in source domain. Our empirical experiments also reveal the limitation of MMD in coping with class weight bias (See Fig.~\ref{fig:IWCB}).

In this paper, we propose a weighted MMD (WMMD) method to address the issue of class weight bias.
As for DA, the challenge is that the class labels in target domain are unknown. So we first introduce class-specific auxiliary weights to reweight the source samples. In this way, the reweighted source data are expected to share the same class weights with target data. The auxiliary weights estimation and model parameters learning are jointly optimized by minimizing the objective function of weighted MMD. Different from MMD, the objective function based on our weighted MMD involves additional weight parameters, and we present a classification EM (CEM) scheme to estimate it. Inspired by the semi-supervised logistic regression in~\cite{r1amini2002semi}, we propose a weighted Domain Adapation Network (WDAN) by both incorporating the weighted MMD into CNN and taking into account the empirical loss on target samples. The CEM algorithm are developed for learning WDAN in three steps,~\ie, E-step, C-step, and M-step.
In the E-step and the C-step, we calculate the class posterior probability, assign the pseudo-labels to the target samples, and estimate the auxiliary weight.
In the M-step, model parameters are updated by minimizing the objective loss. Experimental results show our weighted MMD can learn better domain-invariant representation for domain adaptation. Moreover, the models based on weighted MMD also outperforms the MMD-based counterparts. In summary, the main contributions of this work are three-fold:
\begin{enumerate}
  \item A weighted MMD model is proposed to alleviate the effect of class weight bias in domain adaptation. By taking class prior distributions into account, weighted MMD can provide a better metric for domain discrepancy.
  \item Using unbiased estimate of multi-kernel MMD \cite{gretton2006kernel,gretton2012optimal}, our proposed weighted MMD can be computed as mean embedding matching with linear time complexity and be incorporated into CNN for unsupervised domain adaptation. We further develop a CEM algorithm for training the weighted MMD model.
  \item Experiments demonstrate that weighted MMD outperforms MMD for domain adaptation. The superiority of weighted MMD over MMD has been verified on various CNN architectures and different datasets.
\end{enumerate}

In the remainder of this paper, we begin with a brief introduction to the preliminaries and related work in Section~\ref{section2}. In Section~\ref{section3}, by considering class weight bias, we propose weighted MMD on the basis of conventional MMD. After that, in Section~\ref{section4}, we apply weighted MMD to unsupervised domain adaptation and present a model named WDAN. Extensive experimental results are given in Section~\ref{section5} to verify the effectiveness of our proposed weighted MMD model and detailed empirical analysis to our proposed model is provided. Finally, we conclude this work in Section~\ref{section6}.
\section{Preliminaries and Related Work}\label{section2}
In this section, we first review MMD and its application in domain adaptation, and then survey several other methods used to measure domain discrepancy.
\subsection{MMD and Its Application in Domain Adaptation}\label{subsection2.1}
Domain adaptation aims at adapting the discriminative model learned on source domain into target domain. Depending on the accessibility of class labels for target samples during training, research lines can be grouped into three categories: supervised, semi-supervised, and unsupervised domain adaptation. In this paper, we focus on learning transferable CNN features for unsupervised domain adaptation (UDA), where the labels of all target samples are unknown during training. Compared with the other settings, UDA is more ubiquitous in real-world applications.

Due to the unavailability of labels in the target domain, one commonly used strategy of UDA is to learn domain invariant representation via minimizing the domain distribution discrepancy. Maximum Mean Discrepancy (MMD) is an effective non-parametric metric for comparing the distributions based on two sets of data \cite{borgwardt2006integrating}. Given two distributions $s$ and $t$, by mapping the data to a reproducing kernel Hilbert space (RKHS) using function $\phi(\cdot)$, the MMD between $s$ and $t$ is defined as,
\begin{equation}\label{Eqn1}
\small
{{\rm{MMD}}^2}(s,t) = \mathop {\sup }\limits_{{{\left\| \phi  \right\|}_{\cal H}} \le 1} \left\| {{E_{{{\mathbf{x}}^s} \sim s}}\left[ {\phi ({{\mathbf{x}}^s})} \right] - {E_{{{\mathbf{x}}^t} \sim t}}\left[ {\phi ({{\mathbf{x}}^t})} \right]} \right\|_{\cal H}^2,
\end{equation}
where ${E_{{{\mathbf{x}}^s} \sim s}}\left[  \cdot  \right]$ denotes the expectation with regard to the distribution~\emph{s}, and ${\left\| \phi  \right\|_{\cal H}} \le 1$ defines a set of functions in the unit ball of a RKHS ${\cal H}$. Based on the statistical tests defined by MMD, we have ${\rm{MMD}}(s,t) = 0$ iff $s = t$.
Denote by ${{\cal D}_s} = \{ \mathbf{x}_i^s\} _{i = 1}^M$ and ${{\cal D}_t} = \{ \mathbf{x}_i^t\} _{i = 1}^N$ two sets of samples drawn~\emph{i.i.d.} from the distributions $s$ and $t$, respectively. An empirical estimate of MMD can be given by \cite{gretton2012kernel},
\begin{equation}\label{Eqn2}
\footnotesize
{{\rm{MMD}}^2}({{\cal D}_s},{{\cal D}_t}) = \left\| {\frac{1}{M}\sum\limits_{i = 1}^M {\phi (\mathbf{x}_i^s) - \frac{1}{N}\sum\limits_{j = 1}^N {\phi (\mathbf{x}_j^t)} } } \right\|_{\cal H}^2,
\end{equation}
where $\phi(\cdot)$ denotes the feature map associated with the kernel map $k({\mathbf{x}^s},{\mathbf{x}^t}) = \left\langle {\phi ({\mathbf{x}^s}),\phi ({\mathbf{x}^t})} \right\rangle $. $k({\mathbf{x}^s},{\mathbf{x}^t})$ is usually defined as the convex combination of $L$ basis kernels ${k_l}({\mathbf{x}^s},{\mathbf{x}^t})$ \cite{long2015learning},
\begin{equation}\label{Eqn3}
k({\mathbf{x}^s},{\mathbf{x}^t}) = \sum\limits_{l = 1}^L {{\beta _l}{k_l}({\mathbf{x}^s},{\mathbf{x}^t})} ,{\rm{ s}}{\rm{.t}}{\rm{. }}{\beta _l} \ge 0,\sum\limits_{l = 1}^L {{\beta _l}}  = 1.
\end{equation}

Most existing domain adaptation methods~\cite{long2013transfer, pan2011domain, baktashmotlagh2016distribution, zhong2009cross, tzeng2014deep} are based on the MMD defined in Eqn.~(\ref{Eqn2}) and only linear kernel is adopted for simplicity. Because the formulation of MMD in Eqn.~(\ref{Eqn2}) is based on pairwise similarity and is computed in quadratic time complexity, it is prohibitively time-consuming and unsuitable for using mini-batch stochastic gradient descent (SGD) in CNN-based domain adaptation methods. Gretton \etal~\cite{gretton2012kernel} further suggest an unbiased approximation to ${\rm{MMD}}_l$ with linear complexity. Without loss of generality, by assuming $M = N$, ${\rm{MMD}}_l$ can then be computed as,
\begin{equation}\label{Eqn4}
{\rm{MMD}}_l^2(s,t) = \frac{2}{M}\sum\limits_{i = 1}^{{M \mathord{\left/
 {\vphantom {M 2}} \right.
 \kern-\nulldelimiterspace} 2}} {{h_l}({\mathbf{z}_i})},
\end{equation}
where ${h_l}$ is an operator defined on a quad-tuple ${\mathbf{z}_i} = (\mathbf{x}_{2i - 1}^s,\mathbf{x}_{2i}^s,\mathbf{x}_{2j - 1}^t,\mathbf{x}_{2j}^t)$,
\begin{align}
  {h_l}({\mathbf{z}_i})  = & k(\mathbf{x}_{2i - 1}^s,\mathbf{x}_{2i}^s) + k(\mathbf{x}_{2j - 1}^t,\mathbf{x}_{2j}^t) &\nonumber \\
               & - k(\mathbf{x}_{2i - 1}^s,\mathbf{x}_{2j}^t) - k(\mathbf{x}_{2i}^s,\mathbf{x}_{2j - 1}^t). &\label{Eqn5}
\end{align}
The approximation in Eqn.~(\ref{Eqn4}) takes a summation form and is suitable for gradient computation in a mini-batch manner. Based on the work in~\cite{gretton2012kernel}, Long \etal~\cite{long2015learning} propose deep adaptation networks and residual transfer networks for UDA by introducing ${\rm{MMD}}_l$ based adaptation layers into deep CNNs. However, the existing MMD-based UDA approaches all assume that the source and target data have the same class prior distributions, which does not always hold in real-world applications, as illustrated in Fig.~\ref{fig1}. Our empirical experiments show that class weight bias can result in performance degradation for MMD-based UDA.

\subsection{Metrics for Domain Discrepancy}\label{subsection2.2}
Besides MMD, there are several other metrics for measuring domain discrepancy. Baktashmotlagh \etal~\cite{baktashmotlagh2016distribution} propose a distribution-matching embedding (DME) approach for UDA, where both MMD and the Hellinger distance are adopted to measure the discrepancy between the source and target distributions. Instead of embedding of distributions, discriminative methods such as domain classification~\cite{ganin2014unsupervised} and domain confusion \cite{tzeng2015simultaneous} have also been introduced to learn domain invariant representation. However, class weight bias is also not yet considered in these methods.

Several sample reweighting or selection methods \cite{gong2013connecting,huang2006correcting} are similar to our weighted MMD in spirit, and have been proposed to match the source and target distributions. These methods aim to learn sample-specific weights or select appropriate source samples for target data. Different from them, our proposed weighted MMD alleviates class weight bias by assigning class-specific weights to source data.
\section{Weighted Maximum Mean Discrepancy}\label{section3}
In this section, we will introduce the proposed weighted MMD. Denote by ${p_s}({{\mathbf{x}}^s})$  and ${p_t}({{\mathbf{x}}^t})$ the probability density functions of the source data ${\mathbf{x}}^s$ and the target data ${\mathbf{x}}^t$, $y^s$ and $y^t$ be the class labels of ${\mathbf{x}}^s$ and ${\mathbf{x}}^t$, respectively. Actually, both ${p_s}({{\mathbf{x}}^s})$ and ${p_t} ({{\mathbf{x}}^t})$ can be further represented as the mixtures of class conditional distributions,
\begin{align}
{p_u}({{\mathbf{x}}^u}) &= \sum\limits_{c = 1}^C {{p_u}({y^u} = c){p_u}({{\mathbf{x}}^u}|{y^u} = c)} &\nonumber \\
                    &= \sum\limits_{c = 1}^C {w_c^u{p_u}({{\mathbf{x}}^u}|{y^u} = c)}, u \in \{s,t\}, &\label{Eqn6}
\end{align}
where $w_c^s = {p_s}({y^s} = c)$  and $w_c^t = {p_t}({y^t} = c)$  denote the class prior probability (\ie, class weights) of the source and target samples, respectively, and $C$ denotes the number of classes.

Note that, the difference between the class conditional distributions ${p_s}({{\bf{x}}^s}|{y^s} = c)$ and ${p_t}({{\bf{x}}^t}|{y^t} = c)$ serves as a proper metric of domain discrepancy. However, due to the unavailability of class labels for target data in UDA, the MMD between ${p_s}({{\mathbf{x}}^s})$ and ${p_t}({{\mathbf{x}}^t})$ is usually adopted as a domain discrepancy metric. When $w_c^s = w_c^t$ (c = 1, 2, ..., C), we argue that it is a suitable alternative. Unfortunately, as shown in Fig.~\ref{fig1}, the assumption $w_c^s = w_c^t$ generally does not hold. For this case, MMD cannot cope with class weight bias across domains. We propose to construct a reference source distribution ${p_{s,\alpha }}({{\mathbf{x}}^s})$ for comparing the discrepancy between the source and target domains. Specifically, we require that ${p_{s,\alpha }}({{\bf{x}}^s})$ has the same class weights with the target domain but owns the class conditional distributions in source domain. Let ${\alpha _c} = {{w_c^t} \mathord{\left/{\vphantom {{w_c^t} {w_c^s}}} \right.\kern-\nulldelimiterspace} {w_c^s}}$. In order to eliminate the effect of class weight bias, we define ${p_{s,\alpha }}({{\mathbf{x}}^s})$  as,
\begin{equation}\label{Eqn7}
{p_{s,\alpha }}({{\mathbf{x}}^s}) = \sum\limits_{c = 1}^C {{\alpha_c}w_c^s{p_s}({{\mathbf{x}}^s}|{y^s} = c)}.
\end{equation}
Denote by ${{\cal D}_s} = \{ ({\mathbf{x}}_i^s,y_i^s)\} _{i = 1}^M$ the training set from source domain and ${{\cal D}_t} = \{ {\mathbf{x}}_j^t\} _{j = 1}^N$ the test set from the target domain.
Given the class weights of the target samples, the empirical estimation of weighted MMD ${p_{s,\alpha }}({{\mathbf{x}}^s})$  and ${p_t}({{\mathbf{x}}^t})$ can be given by,

\begin{equation}\label{Eqn8}
\small
{\rm{MMD}}_w^2({{\cal D}_s},{{\cal D}_t}) = \left\| {\frac{1}{{\sum\nolimits_{i = 1}^M {{\alpha _{y_i^s}}} }}\sum\limits_{i = 1}^M {{\alpha _{y_i^s}}\phi(\mathbf{x}_i^s)}  - \frac{1}{N}\sum\limits_{j = 1}^N {\phi(\mathbf{x}_j^t)} } \right\|_{\cal H}^2.
\end{equation}
Assuming $M = N$, the linear time complexity approximation of weighted MMD can then be computed as,

\begin{equation}\label{Eqn9}
{\rm{MMD}}_{l,w}^2({{\cal D}_s},{{\cal D}_t}) = \frac{2}{M}\sum\limits_{i = 1}^{{M \mathord{\left/
 {\vphantom {M 2}} \right.
 \kern-\nulldelimiterspace} 2}} {{h_{l,w}}({\mathbf{z}_i})},
\end{equation}
where ${h_{l,w}}({\mathbf{z}_i})$ is an operator defined on a quad-tuple $\mathbf{z}_i = ({\mathbf{x}}_{2i - 1}^s,\mathbf{x}_{2i}^s,\mathbf{x}_{2j - 1}^t,\mathbf{x}_{2j}^t)$,
\begin{align}
               {h_{l,w}}({\mathbf{z}_i}) & = {\alpha _{y_{2i{\rm{ - }}1}^s}}{\alpha _{y_{2i}^s}}k(\mathbf{x}_{2i - 1}^s,{\mathbf{x}}_{2i}^s) + k(\mathbf{x} _{2i - 1}^t,\mathbf{x}_{2i}^t) &\nonumber \\
                                & - {\alpha _{y_{2i{\rm{ - }}1}^s}}k(\mathbf{x}_{2i - 1}^s,\mathbf{x}_{2j}^t) - {\alpha _{y_{2i}^s}}k(\mathbf{x}_{2i}^s,\mathbf{x}_{2j - 1}^t). &\label{Eqn10}
\end{align}

\section{Weighted Domain Adaptation Network}\label{section4}
By far, we have introduced our weighted MMD for measuring domain discrepancy. But there are two remained issues to be addressed. On one hand, the proposed weighted MMD, similar to MMD, should be incorporated into some classifiers for domain adaptation. On the other hand, the class distribution on target domain is generally unknown during training. In this section, we propose a weighted domain adaptation network (WDAN) model, which is in essential an extension of the semi-supervised logistic regression~\cite{r1amini2002semi} by adding the WMMD term and incorporating with CNN. Meanwhile, we employ the CEM~\cite{r2celeux1992classification} framework and show how we optimize the proposed WDAN without the label information of the target samples.

First, based on the research in~\cite{yosinski2014transferable, long2015learning}, the features gradually become task specific as the layers go toward the top one, resulting in increasing dataset bias for the higher layers of features. Therefore, to generalize CNN for domain adaptation, the weighted MMD-based regularizers are added to the higher layers of CNN. Second, the relationship between semi-supervised learning and domain adaptation has been studied in~\cite{r3sogaard2013semi}. To further exploit the unlabelled data on target domain, we follow the semi-supervised CEM model in~\cite{r1amini2002semi}, leading to the following WDAN model,
\begin{align}
   \mathop {\min }\limits_{\mathbf{W},\{\hat{y}_j\}_{j=1}^N,\mbox{\boldmath$\alpha$}}& \frac{1}{M}\sum\limits_{i = 1}^M {\ell ({\mathbf{x}}_i^s,y_i^s;{\bf{W}})} + \gamma\frac{1}{N}\sum\limits_{j = 1}^N {\ell ({\mathbf{x}}_j^t,\hat{y}_j^t;{\bf{W}})}
    &\nonumber \\
   & +\lambda\sum\limits_{l = {l_1}}^{{l_2}} {{{\rm{MMD}}_{l,w}}({\cal D}_s^l,{\cal D}_t^l)} ,
     & \label{Eqn11}
\end{align}
where $\mathbf{W}$ denotes the model parameters to be learned, and $\{\hat{y}_j\}_{j=1}^N$ represent the assigned labels of target samples. $\lambda$ and $\gamma$ are two tradeoff parameters. ${\cal D}_s^l$ and ${\cal D}_t^l$ denote the \emph{l}-th layer features of the source and target domains, respectively. $w_c^s$ is estimated based on the source data ${\cal D}_s^l$,~\ie, $w_c^s = {{{M_c}} \mathord{\left/{\vphantom {{{M_c}} M}} \right. \kern-\nulldelimiterspace} M}$, where ${M}_c$ is the number of samples of the \emph{c}-th class. The first two terms of Eqn.~\eqref{Eqn11} are the soft-max loss items on the source and target samples, respectively. And the third term is the weighted MMD regularizers for the \emph{$l_1$} $\sim$ \emph{$l_2$}-th layers defined in Eqn.~\eqref{Eqn8}.

Next, we explain the optimization procedure of the proposed WDAN model. Following the CEM algorithm in~\cite{r2celeux1992classification}, the WDAN model is optimized by alternating between: (i) E-step: estimating the class posterior probability of $\{{\mathbf{x}}_j^t\}_{j=1}^N$, (ii) C-step: assigning pseudo-labels $\{\hat{y}_j\}_{j=1}^N$ and estimating auxiliary weights ${\boldmath \alpha}$, (iii) M-step: updating the model parameters ${\mathbf{W}}$.
Given the model parameters $\mathbf{W}$, for each ${\mathbf{x}}_j^t$, we first estimate the class posterior probability based on the output of softmax classifier. The pseudo-label to $\hat{y}_j$ is assigned to ${\mathbf{x}}_j^t$ based on the maximum posterior probability, and the auxiliary weights ${\boldmath \alpha}$ are then estimated based on pseudo-labels. Given $\{\hat{y}_j\}_{j=1}^N$ and ${\boldmath \alpha}$, the conventional backpropagation algorithm is then deployed to update $\mathbf{W}$.
In the following, we give more details on the E-step, C-step, and M-step.

\textbf{E-step}: Fixed $\mathbf{W}$, for each sample ${\mathbf{x}}_j^t$ from target domain, the CNN output to the $c$th class is represented as $g_c(\mathbf{x}_j^t; \mathbf{W})$. Here we simply define the class posterior probability $p({y}_j^t = c |\mathbf{x}_j^t)$ as,
\begin{equation}
p({y}_j^t = c |\mathbf{x}_j^t) = g_c(\mathbf{x}_j^t; \mathbf{W}).
\end{equation}

\textbf{C-step}: With $p({y}_j^t = c |\mathbf{x}_j^t)$, we assign pseudo-label $\hat{y}_j$ to $\mathbf{x}_j^t$ by,
\begin{equation}
\hat{y}_j = \arg \max_{c} p({y}_j^t = c |\mathbf{x}_j^t).
\end{equation}
Let ${\mathbf{1}_c}(\hat{y}_j)$  be an indicator function,
\begin{equation}
{\mathbf{1}_c}(\hat{y}_j) = \left\{ {\begin{array}{*{20}{l}}
{1,{\rm{ if }}\ \ \hat{y}_j = c}\\
{0,{\rm{ otherwise.}}}
\end{array}} \right.
\end{equation}
The class weight $\hat{w}_c^t$  can be estimated by $\hat{w}_c^t = {{\sum\nolimits_j {{\mathbf{1}_c}(\hat{y}_j)} } \mathord{\left/{\vphantom {{\sum\nolimits_j {{1_c}(\hat{y}_j^t)} } N}} \right.\kern-\nulldelimiterspace} N}$, where $N$ is the number of target samples. Then the auxiliary weight can be updated with ${\alpha _c} = {{\hat{w}_c^t} \mathord{\left/{\vphantom {{\hat{w}_c^t} {w_c^s}}} \right.\kern-\nulldelimiterspace} {w_c^s}}$ .

\textbf{M-step}: Fixed \mbox{\boldmath$\alpha$}, the subproblem on $\mathbf{W}$ can be formulated as,
\begin{align}
\mathop {\min }\limits_{\mathbf{W}}\mathcal{L}({\mathbf{W}}) & = \frac{1}{M}\sum\limits_{i = 1}^M {\ell ({\mathbf{x}}_i^s,y_i^s;{\mathbf{W}})} + \gamma\frac{1}{N}\sum\limits_{j = 1}^N {\ell ({\mathbf{x}}_j^t,\hat{y}_j^t;{\bf{W}})} &\nonumber\\
                                                   & + \lambda  \sum\limits_{l = {l_1}}^{{l_2}} {{\rm{MMD}}_{l,w}}({\cal D}_s^l,{\cal D}_t^l). &\label{Eqn12}
\end{align}
Since the gradients of the three terms in Eqn.~\eqref{Eqn12} are computable, $\mathbf{W}$ can be updated with mini-batch SGD. Let ${\mathbf{z}_i} = (\mathbf{x}_{2i - 1}^s,\mathbf{x}_{2i}^s,\mathbf{x}_{2j - 1}^t,\mathbf{x}_{2j}^t)$  be a quad-tuple and  ${\mathbf{z}}_i^l{\rm{ = (}}{\mathbf{z}}_{i,1}^l = {\mathbf{f}}_{2i - 1}^{s,l},{\mathbf{z}}_{i,2}^l = {\mathbf{f}}_{2i}^{s,l},{\mathbf{z}}_{i,3}^l = {\mathbf{f}}_{2i - 1}^{t,l},{\mathbf{z}}_{i,4}^l = {\mathbf{f}}_{2i}^{t,l})$ be the \emph{l}-th layer feature representation of ${{\mathbf{z}}_i}$. Given ${{\mathbf{z}}_i}$, the gradient with respect to the \emph{l}-th layer parameter ${{\mathbf{W}}^l}$ can be written as,

\begin{align}
&\frac{{\partial \mathcal{L}({\mathbf{W}})}}{{\partial {{\mathbf{W}}^l}}} = \frac{1}{2}\sum\limits_{j=1}^2 {\frac{{\partial \ell ({\mathbf{z}}_{i,j}^{},y_{i,j};{\mathbf{W}})}}{{\partial {\mathbf{z}}_{i,j}^l}}} \frac{{\partial {\bf{z}}_{i,j}^l}}{{\partial {{\mathbf{W}}^l}}} &\nonumber\\                                                       &+\frac{\gamma}{2}\sum\limits_{j=3}^4 {\frac{{\partial \ell ({\mathbf{z}}_{i,j}^{},\hat{y}_{i,j};{\mathbf{W}})}}{{\partial {\mathbf{z}}_{i,j}^l}}} \frac{{\partial {\bf{z}}_{i,j}^l}}{{\partial {{\mathbf{W}}^l}}} + \lambda \sum\limits_{k=1}^4 {\frac{{\partial {h_{l,w}}({{\mathbf{z}}_i})}}{{\partial {\mathbf{z}}_{i,k}^l}}\frac{{\partial {\mathbf{z}}_{i,k}^l}}{{\partial {{\mathbf{W}}^l}}}}.&\label{Eqn13}
\end{align}
Taking $k = 1$ for example, $\frac{{\partial {h_{l,w}}({\bf{z}}_i^l)}}{{\partial {\bf{z}}_{i,1}^l}}$ can be computed as,

\begin{align}
\frac{{\partial {h_{l,w}}({{\mathbf{z}}_i})}}{{\partial {\mathbf{f}}_{2i - 1}^{s,l}}}  = & {\alpha _{y_{2i{\rm{ - }}1}^s}}{\alpha _{y_{2i}^s}}\frac{{\partial k({\mathbf{f}}_{2i - 1}^{s,l},{\mathbf{f}}_{2i}^{s,l})}}{{\partial {\mathbf{f}}_{2i - 1}^{s,l}}} &\nonumber\\
                                                       & - {\alpha _{y_{2i{\rm{ - }}1}^s}}\frac{{\partial k({\mathbf{f}}_{2i - 1}^{s,l},{\mathbf{f}}_{2i}^{t,l})}}{{\partial {\mathbf{f}}_{2i - 1}^{s,l}}}. & \label{Eqn14}
\end{align}
Similarly, $\frac{{\partial {h_{l,w}}({\mathbf{z}}_i^l)}}{{\partial {\mathbf{z}}_{i,k}^l}}$ can also be computed for other \emph{k} values. Thus, the model parameters can be updated via backpropagaton with a mini-batch of quad-tuple. Moreover, following~\cite{long2015learning,gretton2012optimal}, the multiple kernel parameters \mbox{\boldmath$\beta$} can also be updated during training.\\

The algorithm described above actually is an extension of classification EM. The C-step in~\cite{r2celeux1992classification} only assigns pseudo-label to each unlabeled sample, while in this work we further estimate the auxiliary weights ${\boldmath \alpha}$ with the pseudo-labels. As shown in~\cite{r2celeux1992classification}, such a optimization procedure can converge to a stationary value. The experiment also empirically validate that our algorithm works well in estimating the auxiliary weights \mbox{\boldmath$\alpha$}.


\begin{table*}[htbp]
\begin{center}
\begin{tabular}{|p{3.5cm}|c|c|c|c|c|c|c|}
\hline
Method & A$\rightarrow$C & W$\rightarrow$C & D$\rightarrow$C & C$\rightarrow$A & C$\rightarrow$W & C$\rightarrow$D & Avg. \\
\hline\hline
AlexNet~\cite{krizhevsky2012imagenet} & 84.0$\pm$0.3 & 77.9$\pm$0.4 & 81.0$\pm$0.4 & 91.3$\pm$0.2 & 83.2$\pm$0.3 & 89.1$\pm$0.2 & 84.0 \\
LapCNN (AlexNet)~\cite{weston2012deep} & 83.6$\pm$0.6 & 77.8$\pm$0.5 & 80.6$\pm$0.4 & 92.1$\pm$0.3 & 81.6$\pm$0.4 & 87.8$\pm$0.4 & 83.9 \\
DDC (AlexNet)~\cite{tzeng2014deep}& 84.3$\pm$0.5 & 76.9$\pm$0.4 & 80.5$\pm$0.2 & 91.3$\pm$0.3 & 85.5$\pm$0.3 & 89.1$\pm$0.3 & 84.6 \\
DAN (AlexNet)~\cite{long2015learning} & 86.0$\pm$0.5 & 81.5$\pm$0.3 & 82.0$\pm$0.4 & 92.0$\pm$0.3 & 92.6$\pm$0.4 & 90.5$\pm$0.2 & 87.3 \\
\hline
WDAN (AlexNet) &\textbf{86.9}$\pm$0.1& \textbf{84.1}$\pm$0.2 & \textbf{83.9}$\pm$0.1 & \textbf{93.1}$\pm$0.2 & \textbf{93.6}$\pm$0.2 & \textbf{93.4}$\pm$0.2 & \textbf{89.2} \\
WDAN$^{\star}$ (AlexNet) & 87.1$\pm$0.2 & 85.1$\pm$0.3 & 85.2$\pm$0.2 & 93.2$\pm$0.1 & 93.5$\pm$0.3 & 94.5$\pm$0.2 & 89.8 \\
\hline\hline
GoogLeNet~\cite{szegedy2015going} & 91.3$\pm$0.2 & 88.2$\pm$0.3 & 88.9$\pm$0.3 & 95.2$\pm$0.1 & 92.5$\pm$0.2 & 94.7$\pm$0.3 & 91.8  \\
DDC (GoogLeNet)~\cite{tzeng2014deep} & 91.4$\pm$0.2 & 88.7$\pm$0.3 & 89.0$\pm$0.4 & 95.3$\pm$0.2 & 93.0$\pm$0.1 & 94.9$\pm$0.4 & 92.1 \\
DAN (GoogLeNet)~\cite{long2015learning} & 91.4$\pm$0.3 & 89.7$\pm$0.2 & 89.1$\pm$0.4 & 95.5$\pm$0.2 & 93.1$\pm$0.3 & 95.3$\pm$0.1 & 92.3 \\
\hline
WDAN (GoogLeNet) & \textbf{92.2}$\pm$0.2 & \textbf{91.0}$\pm$0.5 & \textbf{89.8}$\pm$0.3 & \textbf{95.5}$\pm$0.3 & \textbf{95.4}$\pm$0.2 & \textbf{95.5}$\pm$0.5 & \textbf{93.2} \\
\hline\hline
VGGnet-16~\cite{simonyan2014very} & 89.6$\pm$0.4 & 88.1$\pm$0.4 & 85.4$\pm$0.5 & 93.7$\pm$0.2 & 94.3$\pm$0.2 & 93.7$\pm$0.2 & 90.8  \\
DAN (VGGnet-16)~\cite{long2015learning} & 91.2$\pm$0.2 & 90.6$\pm$0.3 & 87.1$\pm$0.4 & 95.7$\pm$0.2 & 95.3$\pm$0.3 & 94.7$\pm$0.1 & 92.4 \\
\hline
WDAN (VGGnet-16) & \textbf{91.4}$\pm$0.2 & \textbf{91.0}$\pm$0.2 & \textbf{89.0}$\pm$0.3 & \textbf{95.7}$\pm$0.1 & \textbf{95.8}$\pm$0.2 & \textbf{95.9}$\pm$0.3 & \textbf{93.1} \\
\hline
\end{tabular}
\end{center}
\caption{Results (in $\%$) of different methods based on \emph{AlexNet}, \emph{GoogleNet} and \emph{VGGnet-16} on \emph{Office-10+Caltech-10}. Note that the results of LapCNN, DDC and DAN are duplicated from \cite{long2015learning}. $\star$ indicates that the ground truth class distributions in both source and target domain are used as prior.}
\label{office10}
\end{table*}
\section{Experiments}\label{section5}
In this section, we first evaluate our proposed WDAN on four widely used benchmarks in UDA, \ie, \emph{Office-10+Caltech-10}~\cite{gong2012geodesic},~\emph{Office31}~\cite{saenko2010adapting},~\emph{ImageCLEF}~\cite{caputo2014overview} and \emph{Digit Recognition}. Moreover, we also provide empirical analysis to our proposed WDAN model from three aspects,~\ie, hyper-parameter sensitivity, robustness to class weight bias, and feature visualization.

Following the common setting in UDA, we implement our WDAN model based on four widely used CNN architectures,~\ie, LeNet~\cite{Lecun1998GradientbasedLA}, AlexNet~\cite{krizhevsky2012imagenet}, GoogLeNet~\cite{szegedy2015going} and VGGnet-16~\cite{simonyan2014very}. As suggested in ~\cite{long2015learning}, we train our method based on pre-trained AlexNet, VGGNet-16, or GoogLeNet on ImageNet, with the layers from conv1 to conv3 fixed for AlexNet and inception layers from inc1 to inc3 fixed for GoogLeNet. The WDAN (LeNet) is trained from the scratch (random initialization). In addition, the auxiliary weight is initialized with $\alpha_c = 1$ for each class. For $l_1$ and $l_2$, we follow the setting in~\cite{long2015learning}. Concretely, WMMD-based regularizers are added to the last three fully connected layers for AlexNet, the last inception and fully connected layers for GoogleNet, and the last fully connected layer for LeNet. All experiments are implemented by using Caffe Toolbox~\cite{jia2014caffe}, and run on a PC equipped with a NVIDIA GTX 1080 GPU and 32G RAM. We set the batch size to 64 for all methods, and optimize the learning rate for each model independently. The tradeoff parameters $\lambda$ and $\gamma$ are optimized in sets $\{0,~0.03,~0.07,~0.1,~0.4,~0.7,~1.4,~1.7,~2\}$ and  $\{0,~0.1,~0.2,~0.3,~0.4,~0.5,~0.6,~0.7,~0.8,~0.9,~1.0\}$ by cross-validation, respectively. The source code of our WDAN is available at: \url{https://github.com/yhldhit/WMMD-Caffe}.
\subsection{Comparison with State-of-the-arts}
For UDA, we employ the standard protocols as~\cite{long2015learning,li2015generative,long2016unsupervised}, where all the samples in source and target domain are used for training. The averaged results over 10 trials on target domain set are reported for comparison.
\subsubsection{Office-10+Caltech-10}
\emph{Office-10+Caltech-10}~\cite{gong2012geodesic} is widely used for domain adaptation, which picks up 10 classes shared in \emph{Office-31}~\cite{saenko2010adapting} and~\emph{Caltech-256}~\cite{griffin2007caltech}. It consists of four domains where Amazon (A), Webcam (W) and  DSLR (D) are from \emph{Office-31}, and the another one is \emph{Caltech-256} (C).
On this dataset, we conduct experiments based on AlexNet, GoogLeNet and VGGnet-16, and exploit the same setting as~\cite{long2015learning} for performance comparison.

We compare our WDAN with several state-of-the-art methods as listed in Table~\ref{office10}, including its MMD counterpart DAN \cite{long2015learning}. By AlexNet, GoogLeNet, and VGGnet-16 we indicate to fine-tune the pre-trained CNN models for special tasks. LapCNN~\cite{weston2012deep} can be seen as a variant of CNN, which first shows deep structure learning can be improved by jointly learning an embedding with the unlabeled data, and then exploits the embedding as a regularizer. By embedding a single kernel MMD layer into CNN structure, DDC~\cite{tzeng2014deep} develops a unified deep framework to jointly learn semantically meaningful feature and perform adaption cross domain.

Numerical results in Table~\ref{office10} show that our weighted DAN achieves the best performance, independently of the employed CNN structure. Moreover, the WDAN is superior to DAN by $1.9\%$, $0.9\%$ and $0.7\%$, respectively. We contribute this improvement to that our weighted MMD model can alleviate the effect of class weight bias. In addition, the superiority over other state-of-the-art methods demonstrate the effectiveness of the proposed WDAN. Finally, we exploit the ground truth class distributions in both source and target domains as prior of the proposed WDAN based on AlexNet, which is indicated as WDAN$^{\star}$ (AlexNet) in Table~\ref{office10}. Although WDAN$^{\star}$ can further improve the performance of WDAN, the smaller gap between them than one between weighted DAN and DAN validate the effectiveness of our proposed learning and estimation method.

\subsubsection{ImageCLEF}
\emph{ImageCLEF}~\cite{caputo2014overview} is developed for the ImageCLEF domain adaptation task\footnote{\url{http://www.imageclef.org/2014/adaptation}}. This dataset collects images from five widely used image benchmarks, including \emph{Caltech256}~\cite{griffin2007caltech}, \emph{Bing}, \emph{PASCAL VOC2012}~\cite{everingham2010pascal}, \emph{ImageNet2012}~\cite{deng2009imagenet} and \emph{SUN}~\cite{xiao2010sun}. This dataset is thought to be more difficult, since some domains contain low-quality images, making this benchmark a good compliance to the \emph{Office-10+Caltech-10}, where the domain is more similar. Different from the original experimental setting, in this paper, we use a subset of \emph{ImageCLEF}, which contains three datasets, \ie, \emph{Caltech256} (C), \emph{Bing} (B) and \emph{PASCAL VOC2012} (P). Meanwhile, we exploit all images in each subset rather than follow the standard protocol to sample the same number of images for each class. Such setting results in six domain adaptation tasks.
\begin{table}[t]
\tiny
\begin{center}
\begin{tabular}{|l|c|c|c|c|c|c|c|}
\hline
Method & P$\rightarrow$C & B$\rightarrow$C & C$\rightarrow$B & P$\rightarrow$B & C$\rightarrow$P & B$\rightarrow$P & Avg. \\
\hline\hline
GoogLeNet & 91.0$\pm$0.5 & 92.4$\pm$0.3 & 61.2$\pm$0.4 & \textbf{55.3}$\pm$0.3 & 61.2$\pm$0.2 & 58.1$\pm$0.3 & 69.9 \\
DDC~\cite{tzeng2014deep}& 91.2$\pm$0.4 & 92.6$\pm$0.4 & 62.0$\pm$0.3 & 54.3$\pm$0.3 & 61.7$\pm$0.4 & 58.6$\pm$0.3 & 70.1 \\
DAN & 91.4$\pm$0.2 & 93.0$\pm$0.1 & 62.5$\pm$0.3 & 54.5$\pm$0.2 & 62.2$\pm$0.3 & 59.0$\pm$0.3 & 70.4 \\
\hline
WDAN & \textbf{91.4}$\pm$0.1 & \textbf{93.8}$\pm$0.2 & \textbf{62.9}$\pm$0.3 & 55.2$\pm$0.3 & \textbf{65.0}$\pm$0.2 & \textbf{59.5}$\pm$0.3 & \textbf{71.3} \\
\hline
\end{tabular}
\end{center}
\caption{Results (in $\%$) of different methods based on \emph{GoogLeNet} on \emph{ImageCLEF} dataset.}
\label{tab_imageclef}
\end{table}

\begin{table}[t]
\footnotesize 
\begin{center}
\begin{tabular}{|l|c|c|c|c|c|c|c|}
\hline
Method & M$\rightarrow$S & S$\rightarrow$M & M$\rightarrow$U & U$\rightarrow$M & Avg \\
\hline\hline
LeNet & 17.2$\pm$0.3 & 56.8$\pm$0.5 & 61.5$\pm$0.4 & 46.5$\pm$0.6 & 45.5 \\
SA~\cite{fernando2013unsupervised}& 21.1$\pm$0.2 & 59.3$\pm$0.3 & 55.0$\pm$0.4 & 51.6$\pm$0.6 & 46.8\\
DAN & 19.3$\pm$0.4 & 65.2$\pm$0.3 & 69.1$\pm$0.5 & 60.5$\pm$0.7 & 53.5 \\
\hline
WDAN & \textbf{23.4}$\pm$0.2 & \textbf{67.4}$\pm$0.4 & \textbf{72.6}$\pm$0.3 & \textbf{65.4}$\pm$0.4 & \textbf{57.2} \\
\hline
\end{tabular}
\end{center}
\caption{Results (in $\%$) of different methods based on \emph{LeNet} on \emph{Digit Classification}.}
\label{DigRec}
\end{table}

\begin{table}[t] \addtolength{\tabcolsep}{-0.5pt}
\setlength{\abovecaptionskip}{0pt}
\setlength{\belowcaptionskip}{0pt}
\tiny
\begin{center}
\begin{tabular}{|l|c|c|c|c|c|c|c|}
\hline
Method & A$\rightarrow$W & D$\rightarrow$W & W$\rightarrow$D & A$\rightarrow$D & D$\rightarrow$A & W$\rightarrow$A & Avg. \\
\hline\hline
AlexNet & 60.4$\pm$0.5 & 94.0$\pm$0.3 & 92.2$\pm$0.3 & 58.6$\pm$0.6 & 46.0$\pm$0.6 & 49.0$\pm$0.5 & 66.7 \\
DAN & 66.0$\pm$0.5 & 94.3$\pm$0.3 & 95.2$\pm$0.3 & 63.2$\pm$0.4 & 50.0$\pm$0.5 & 51.1$\pm$0.5 & 70.0 \\
\hline
WDAN & \textbf{66.8}$\pm$0.3 & \textbf{95.9}$\pm$0.1 & \textbf{98.7}$\pm$0.4 & \textbf{64.5}$\pm$0.2 & \textbf{53.8}$\pm$0.1 & \textbf{52.7}$\pm$0.2 & \textbf{72.1} \\
\hline
\end{tabular}
\end{center}
\vspace{-.1in}
\caption{Results (in $\%$) of different methods based on \emph{AlexNet} on \emph{Office-31} dataset.}
\label{office-31}
\vspace{-.15in}
\end{table}

We compare WDAN with three related methods based on GoogLeNet, \ie, GoogLeNet, DDC and DAN. We implement them by using the codes released by authors\footnote{\url{https://github.com/longmingsheng/mmd-caffe}}, and try our best to optimize them. The results of the competing methods are shown in Table~\ref{tab_imageclef}, from which we can see that our proposed weighted DAN obtains the best performance in most of the cases, and achieves $1.4\%$, $1.2\%$ and $0.9\%$ gains over GoogLeNet, DDC and DAN on average, respectively. The above results show the proposed weighted MMD is helpful to improve the performance of domain adaptation task.

\subsubsection{Digit Recognition}
Furthermore, we conduct experiment on digit recognition, which is usually adopted to domain adaptation. In this paper, we only considering training images of three benchmarks, \ie, \emph{MNIST} (M), \emph{SVHN} (S) and \emph{USPS} (U) and conduct experiments on four tasks. As LeNet~\cite{Lecun1998GradientbasedLA} is usually used for digit recognition, we implement our WDAN and the competing methods based on it. Among them, SA~\cite{fernando2013unsupervised} proposes a subspace alignment method for domain adaptation, which aims at learning a feature mapping to align source samples with target samples. For fair comparison, we implement SA by using the features from the fine-tuned LeNet. The results reported in Table~\ref{DigRec} clearly show that our proposed WDAN achieves the best performance on all tasks, and outperforms LeNet, SA and DAN by $11.7\%$, $10.4\%$ and $3.7\%$ on average, respectively. The significant improvements over competing methods show the proposed weighted MMD model is effective and meaningful.

\begin{figure}[t]
\begin{center}
\includegraphics[width=0.85\linewidth, height = 4.4 cm]{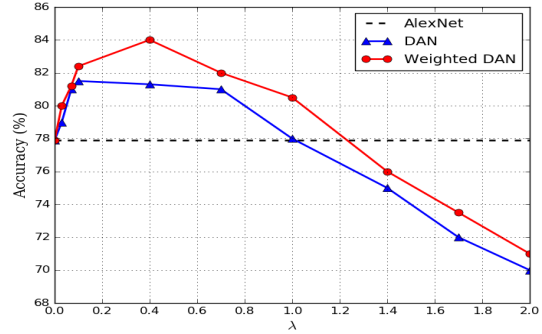}
\end{center}
\caption{Performance (in $\%$) of different methods \emph{w.r.t.} $\lambda$.}
\label{fig:lambda}
\end{figure}

\subsubsection{Office-31}
Finally, experiments are further conducted to assess WDAN on datasets with more classes. We conduct experiments on a dataset with 31 classes (\ie Office-31).
There are three domains, \ie, Amazon (A), Webcam (W) and DSLR (D), in Office-31. In this part, we consider all the six UDA tasks, and report the results using Alexnet.
Table~\ref{office-31} shows the results of AlexNet, DAN, and WDAN.
It can be seen that the proposed WMMD achieves better results than its MMD counterpart, indicating that WMMD also works well on dataset with more classes.
To sum up, the promising performance of our weighted MMD model can be verified on various CNN architectures (\ie, AlexNet, GoogLeNet and LeNet) and various datasets with different number of classes.


\subsection{Empirical Analysis}
In this subsection, we perform empirical analysis of the proposed WDAN from three aspects. Firstly, we evaluate the effect of hyper-parameter $\lambda$ on our proposed WDAN model in Eqn.~(\ref{Eqn11}). Secondly, compared with its baselines, \ie, Alexnet and DAN, we show our proposed WDAN is robust to class weight bias. Finally, we make a visualization of learned feature representations.
\subsubsection{Effect of Parameter $\lambda$}
The objective in Eqn.~(\ref{Eqn11}) of WDAN consists of three terms, \ie, conventional empirical losses on the source and target domains, and MMD-based regularizer. Generally speaking, the empirical risk term keeps the learned deep feature to be discriminative on source domain while the MMD-based regularizer encourages domain invariant feature representation. Both of this two aspects are of essential importance for domain adaptation. The parameter $\lambda$ in the objective Eqn.~(\ref{Eqn11}) makes a tradeoff between this two parts, and could greatly impact the performance of domain adaptation. To have a closer look at this parameter, we evaluate our proposed WDAN based on AlexNet on the task W$\rightarrow$C from \emph{Office-10+Caltech-10} under various $\lambda$. As suggested above, $\lambda$ belongs to the set \{0.0,~0.03,~0.07,~0.1,~0.4,~0.7,~1,~1.4,~1.7,~2\}. Meanwhile, we also compared our WDAN model with DAN under various $\lambda$. AlexNet is reported as baseline and corresponds to the case $\lambda=0$. The results are illustrated in Fig.~\ref{fig:lambda}.

Obvious conclusions can be drawn from the results: (i) our proposed WDAN consistently outperforms the DAN, demonstrating that mining the class weight bias in MMD is meaningful and beneficial; (ii) WDAN and DAN achieve the best results at  $\lambda=0.4$ and  $\lambda=0.1$, and outperforms the baseline, \ie, AlexNet, when $\lambda<1.2$ and  $\lambda<1.0$, respectively, indicating that an appropriate balance is important and necessary.

\subsubsection{Impact of Class Weight Bias}\label{subsection522}
To further clarify the impact of class weight bias on MMD-based domain adaptation methods, we conduct experiments on a variant of the task V$\rightarrow$C from \emph{ImageCLEF} based on AlexNet. Specifically, we pick up two shared classes, \ie, airplane and motorbike, in the source domain \emph{PASCAL VOC2012} (V) and target domain \emph{Caltech256} (C), which forms a two-class classification problem.

\begin{figure}[t]
\begin{center}
\includegraphics[width=0.85\linewidth, height = 4.4 cm]{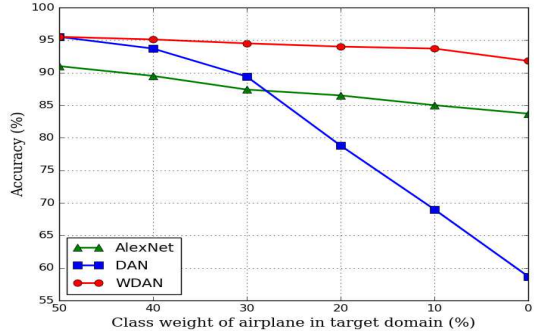}
\end{center}
   \caption{Performance (in $\%$) of different methods \emph{w.r.t.} class weight bias.}
\label{fig:IWCB}
\end{figure}

Then we fix the class weights as 0.5 for each class on source domain and train different methods with gradually changing the class distribution on target domain, which can be interpreted as different level of the class weight bias cross source and target domains. Fig.~\ref{fig:IWCB} show the results of WDAN, DAN and AlexNet under different levels of the class weighted bias.
From it we can see that the class weight bias has great influence on performance of MMD-based domain adaptation methods. Moreover, the conventional MMD-based methods (\eg, DAN) are limited in handling the class weight bias, as its results significantly degrade with increasing class weighted bias. In addition, our proposed WDAN is more robust to class weighted bias.

\subsubsection{Feature Visualization}
\begin{figure}[t]
\begin{center}
\includegraphics[width=1.0\linewidth]{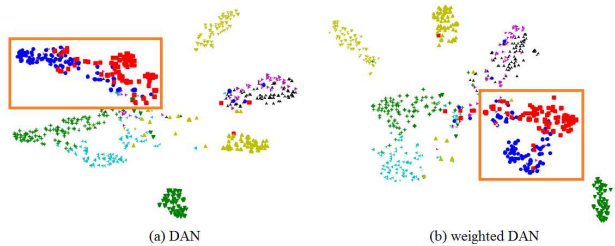}
\end{center}
\caption{The t-SNE visualization of learned features of different methods.}
\label{fig:FV}
\end{figure}

Following the work in~\cite{long2015learning}, we visualize the features learned by WDAN and DAN on target domain in the D$\rightarrow$C task from \emph{Office-10+Caltech-10}. For feature visualization, we employ the t-SNE visualization method~\cite{van2014accelerating} whose source codes are provided\footnote{\url{https://lvdmaaten.github.io/tsne/}}. The results of feature visualization for DAN and weighted DAN are illustrated in Fig.~\ref{fig:FV} (a) and Fig.~\ref{fig:FV} (b), respectively. As shown in the orange boxes of Fig.~\ref{fig:FV}, features learned by the proposed WDAN can reserve more class discrepancy distance than ones learned by DAN. The underlying reason lies in the fact that WDAN, by considering a weighted MMD regularizer, does not minimize the class weight bias as DAN does, which also accounts for that weighted DAN can outperform DAN on a variety of unsupervised domain adaptation tasks.

\section{Conclusion}\label{section6}
In this paper, we focus on the uninvestigated issue of class weight bias in UDA, which has adverse effect on MMD-based domain adaptation methods. We first propose a novel weighted MMD to reduce the effect of class weight bias by constructing a reference source distribution based on target distribution. For UDA, we present a weighted DAN (WDAN) based on the proposed weighted MMD, and develop modified the CEM learning algorithm to jointly assign pseudo-labels, estimate the auxiliary weights and learn model parameters. Empirical results show that our proposed WDAN outperforms its MMD counterpart, \ie, DAN, in various domain adaptation tasks. In future, there remains several issues to be investigated: (i) evaluation of weighted MMD on non-CNN based UDA models, (ii) applications to other tasks (\eg, image generation) based on measuring the discrepancy between distributions.
\section{Acknowledgment}\label{section7}
This work is supported in part by NSFC grant (61671182, 61471082, and 61370163). The authors also thank NVIDIA corporation for the donation of GTX 1080 GPU.
{\small
\bibliographystyle{ieee}
\bibliography{egbib}
}

\end{document}